\documentclass[11pt]{article}

\usepackage[preprint]{acl}

\usepackage{times}
\usepackage{latexsym}

\usepackage[T1]{fontenc}

\usepackage[utf8]{inputenc}
\ifdefined\XeTeXversion
\usepackage{newunicodechar}
\newunicodechar{’}{\textquoteright}
\newunicodechar{–}{\textendash}
\fi

\usepackage{microtype}

\usepackage{inconsolata}

\usepackage{graphicx}
\usepackage{array}
\usepackage{booktabs}
\usepackage{amsmath}
\usepackage{xspace}
\usepackage{comment}
\usepackage{multirow}
\usepackage{xurl}
\usepackage{xspace}
\usepackage{graphicx}
\usepackage{multirow}
\usepackage{tabularx}
\usepackage{tcolorbox}
\usepackage{amsmath}
\usepackage{verbatim}
\usepackage{float}      %
\usepackage{adjustbox}  %
\tcbuselibrary{breakable}
\newtcolorbox{promptbox}[1]{breakable,colback=gray!4,colframe=gray!55,
  title={#1},fonttitle=\bfseries\small,fontupper=\small}

\newcommand{\tool}{Q-Reflector\xspace}
\usepackage{enumitem}
\usepackage{listings}
\ifdefined
\else
\hypersetup{pdfinfo={TemplateVersion={2027.1}}}
\fi

\title{\tool: Learning GUI Action Reflection\\with Generated Conditions}

\author{
  Linqiang Guo\textsuperscript{1,2}\hspace{0.65em}%
  Wei Liu\textsuperscript{1}\hspace{0.65em}%
  Li Gu\textsuperscript{1,2}\hspace{0.65em}%
  Yang Wang\textsuperscript{1,2}\hspace{0.65em}%
  Tse-Hsun (Peter) Chen\textsuperscript{1,2} \\
  \textsuperscript{1}Concordia University \quad \textsuperscript{2}Mila -- Qu\'ebec AI Institute
}

\usepackage{fancyhdr}
\definecolor{QRAccent}{HTML}{7552AA}
\definecolor{QRRule}{HTML}{CFC4E0}
\definecolor{QRAbstract}{HTML}{F6F3FA}
\fancypagestyle{qreflector}{%
  \fancyhf{}%
  \fancyhead[L]{\sffamily\small\bfseries Q-Reflector}%
  \fancyhead[R]{\sffamily\small\thepage}%
  \renewcommand{\headrulewidth}{0.4pt}%
  \renewcommand{\headrule}{\hbox to\headwidth{\color{QRRule}\leaders\hrule height \headrulewidth\hfill}}%
}
\fancypagestyle{qreflectorfirst}{%
  \fancyhf{}%
  \fancyhead[L]{\sffamily\large\bfseries Q-Reflector}%
  \renewcommand{\headrulewidth}{0.5pt}%
  \renewcommand{\headrule}{\hbox to\headwidth{\color{QRRule}\leaders\hrule height \headrulewidth\hfill}}%
}
\makeatletter
\newcommand{\makearxivtitle}{%
  \begingroup
  \centering
  {\fontsize{20}{23}\selectfont\bfseries\@title\par}%
  \vspace{12pt}%
  {\normalsize\outauthor\par}%
  \vspace{11pt}%
  \endgroup
}
\makeatother
\newenvironment{arxivabstract}{%
  \begin{tcolorbox}[colback=QRAbstract,colframe=QRRule,boxrule=0.4pt,
    arc=2pt,left=10pt,right=10pt,top=7pt,bottom=8pt,
    before skip=0pt,after skip=12pt,
    fontupper=\fontsize{10.5}{12.8}\selectfont]
  {\centering\large\bfseries Abstract\par}\vspace{3pt}
  \noindent\ignorespaces
}{\par\end{tcolorbox}}

\hypersetup{
  pdftitle={Q-Reflector: Learning GUI Action Reflection with Generated Conditions},
  pdfauthor={Linqiang Guo, Wei Liu, Li Gu, Yang Wang, Tse-Hsun (Peter) Chen}
}

\begin{document}
\onecolumn
\pagestyle{qreflector}
\thispagestyle{qreflectorfirst}
\makearxivtitle
\begin{arxivabstract}
Mobile GUI agents require accurate action reflection for reliable long-horizon execution, assessing whether each action was successfully executed and whether its outcome constitutes a valid step toward the task goal.
When evaluated as standalone reflectors under a common prompt, frontier VLMs and specialized GUI models exhibit imbalanced recognition of valid and invalid transitions.
We propose \tool\footnotemark, a standalone reflector based on Qwen3-VL-8B that learns GUI action reflection as supervised structured prediction.
Given task context, the executed action, and paired screenshots, it generates pre- and post-conditions specifying the action's prerequisites and intended local effect, then assesses execution success and task validity through a structured rationale and binary verdict.
Training combines supervised fine-tuning, teacher--student distillation, and reward- and preference-based refinement.
On 1,082 manually verified AndroidWorld transitions, \tool achieves 82.16\% overall accuracy, exceeding all evaluated baselines and showing more balanced performance than GPT-5.2 and GUI-Owl-1.5-8B.
On 117 MobileWorld tasks, \tool achieves task success rates of 46.15\% with GUI-Owl-1.5-8B-Instruct and 35.90\% with MAI-UI-8B, improving over no external reflection by 17.09 and 14.53 percentage points, respectively.
Under the same condition-generation setting, it outperforms the strongest reflection baselines, GPT-5.2 and Claude Haiku 4.5, by 3.42 and 5.98 percentage points, respectively.

\end{arxivabstract}
\footnotetext{Code is publicly available at \url{https://github.com/pockyitachi/Q-Reflector}.}

\begingroup
\centering
\includegraphics[width=\textwidth]{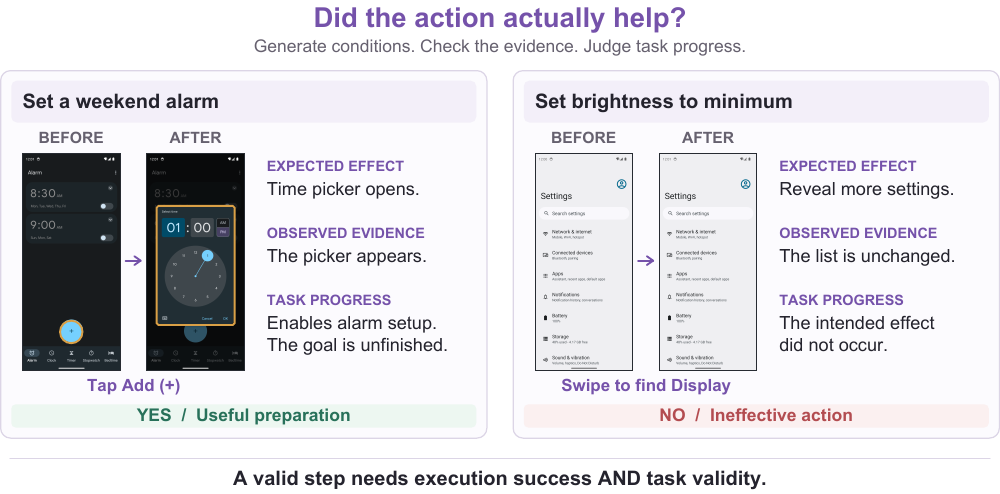}
\captionsetup{font=small,labelfont=bf,skip=5pt,hypcap=false}
\captionof{figure}{\textbf{What should action reflection recognize?}
Opening the time picker is useful preparation, even though the alarm is not yet set.
A swipe that leaves Settings unchanged fails to achieve its intended local effect.
\tool generates conditions and checks execution success and task validity.
Screenshots are from recorded trajectories; conditions and judgments are illustrative.}
\label{fig:teaser}
\par\endgroup

\twocolumn

\section{Introduction}
Mobile GUI agents built on vision--language models (VLMs) can navigate applications and carry out multi-step workflows through visual perception and low-level interaction~\citep{zhang2023appagent,wang2025mobileagente}.
Reliable execution also requires judging whether an action succeeds and whether its resulting transition constitutes a valid step toward the task goal.
Action reflection guides the agent's decision to continue, retry, recover, or stop.
Accepting an invalid transition can allow execution to continue despite a failed or task-irrelevant action, while rejecting a valid transition can trigger unnecessary recovery or interrupt useful progress. When reflection runs as a separate call after each action, its inference cost accumulates over a long task~\citep{li2025mobileuse}.

Figure~\ref{fig:teaser} contrasts useful preparation with an ineffective action: a valid step need not finish the task, but it must achieve a useful local effect.

Action reflection requires more than identifying changes between screenshots.
A visible change may result from a successfully executed but task-irrelevant action.
Conversely, a valid intermediate step may prepare for a later action or recover from an error without completing the task.
Opening a date picker, for example, is a valid preparatory step in booking an appointment.
The reflector must determine what local effect the action should produce, whether that effect occurred, and whether it constitutes a valid task step.

\begingroup\frenchspacing
\fussy\tolerance=1000\emergencystretch=1em\hyphenpenalty=30
\looseness=-1
\widowpenalty=10000
Existing approaches include prompted reflection modules~\citep{wang2025mobileagente,guo2025agentsama} and GUI models trained to verify action outcomes and recover from execution errors~\citep{wu2025guireflection,liu2025infiguiagent}.
GUI-Owl-1.5's training pipeline, for example, synthesizes reflection on action outcomes to guide subsequent action generation~\citep{xu2026mobileagentv35}.
When evaluated as standalone reflectors under a common prompt, both frontier VLMs and specialized GUI models exhibit imbalanced recognition of valid and invalid transitions.
On our 1,082 manually verified AndroidWorld transitions~\citep{rawles2024androidworld}, GPT-5.2 achieves 95.10\% negative-class accuracy but only 51.39\% positive-class accuracy. Under the same reflection prompt, GUI-Owl-1.5-8B shows the opposite pattern, with 92.33\% positive-class accuracy and 57.78\% negative-class accuracy. Reliable reflection therefore requires both accepting valid intermediate steps and rejecting failed or task-invalid~ones.
\par\endgroup

We formulate GUI action reflection as supervised structured prediction.
The reflector first generates action-specific preconditions and postconditions, specifying the action's prerequisites and intended local effect.
It is instructed to derive these conditions from the task context, action, recent history, and screenshot before execution.
It then checks the observed transition against the conditions and separately assesses task validity through a structured rationale before producing a binary decision.
Training supervises the conditions, rationale, and decision together.

We introduce \textbf{\tool}, a compact standalone reflector based on Qwen3-VL-8B~\citep{bai2025qwen3vltechnicalreport}.
Given the task goal, current action, recent action history, and paired screenshots, it generates the conditions, rationale, and decision within a single post-action call.
When a transition is judged invalid, the action-generating model receives feedback extracted from the rationale.
We train \tool on 880 manually reviewed transitions collected by Agent-SAMA from Mobile-Eval-E and SPA-Bench~\citep{wang2025mobileagente,chen2025spabench}, using supervised fine-tuning, teacher--student distillation, and reward- and preference-based refinement.
\tool is added to the execution loop as a separate reflection module, with the action-generating model kept fixed.

On the AndroidWorld transitions, \tool achieves 82.16\% overall accuracy, exceeding GPT-5.2 by 11.83 percentage points given the same transition information.
Its positive- and negative-class accuracies are 86.79\% and 76.12\%, respectively.
These results show more balanced performance across positive and negative transitions than GPT-5.2 and GUI-Owl-1.5-8B.
We further evaluate \tool on 117 MobileWorld tasks~\citep{kong2025mobileworldbenchmarkingautonomousmobile} using GUI-Owl-1.5-8B-Instruct~\citep{xu2026mobileagentv35} and MAI-UI-8B~\citep{zhou2025maiuitechnicalreportrealworld} as action-generating models.
\tool achieves task success rates of 46.15\% and 35.90\%, improving over no external reflection by 17.09 and 14.53 percentage points, respectively.
Under the same condition-generation setting, it outperforms the strongest reflection baselines, GPT-5.2 and Claude Haiku 4.5, by 3.42 and 5.98 percentage points, respectively.
Compared with a separately trained variant using the same data and training pipeline but omitting condition generation, \tool completes five additional tasks with GUI-Owl-1.5-8B-Instruct and seven with MAI-UI-8B.

\tool combines more accurate and balanced judgments learned through task-specific training with explicit pre- and post-condition generation that further improves task execution.
Our contributions are as follows:
\begingroup\frenchspacing
\fussy\tolerance=1000\emergencystretch=1em\hyphenpenalty=30
\begin{itemize}[topsep=2pt, itemsep=2pt, parsep=2pt, beginpenalty=10000]
    \item \textbf{Supervised structured reflection.}\newline
    We address the imbalance between accepting valid transitions and rejecting invalid ones through supervised structured prediction, achieving higher overall accuracy and more balanced performance than the GPT-based baselines.
    The model first generates action-specific pre- and post-conditions, then assesses execution success and task validity.
    \item \textbf{A compact, reusable learned reflector.}\newline
    We instantiate this formulation in \tool, an 8B reflection module that generates pre- and post-conditions, a structured rationale, and a binary judgment from task context and paired visual evidence.
    \item \textbf{Static and online evaluation.}\newline
    We evaluate \tool on static AndroidWorld transitions and within MAI-UI-8B and GUI-Owl-1.5-8B-Instruct on MobileWorld, analyzing transition accuracy, end-to-end task success, and per-task token usage.
\end{itemize}
\endgroup

\section{Related Work}
\label{sec:related}

\begingroup
\brokenpenalty=10000

\textbf{Condition Checking and Execution Feedback.}\@
Condition checking and execution feedback support embodied action planning.
ProgPrompt~\citep{singh2023progprompt} generates robot programs with precondition assertions and recovery actions.
REFLECT~\citep{liu2023reflect} summarizes multisensory execution histories to explain failures and guide correction.
\tool learns a dedicated GUI reflector that generates action-specific conditions before checking observed outcomes.

\textbf{Specialized GUI Models.}\@
GUI models have been developed through trajectory generation and reinforcement learning~\citep{ye2025mobileagentv3fundamentalagentsgui,lu2025uir1}, screenshot-based grounding~\citep{cheng2024seeclickharnessingguigrounding,du2025testtimerlgui}, and pre-execution criticism~\citep{wanyan2025lookleapguicriticr1model}.
Agent-SAMA~\citep{guo2025agentsama} uses a prompted Reflection Agent to compare expected and observed states within an FSM; its reflection interface informs our training-data construction.
InfiGUIAgent~\citep{liu2025infiguiagent} and VeriGUI~\citep{zhang2026verigui} learn to generate expected action effects and verify them at the next step within their action policies.
\tool trains reflection separately from action generation, using generated conditions to assess completed GUI transitions.

\begingroup\frenchspacing
\textbf{Reflection for GUI Agents.}\@
MobileUse~\citep{li2025mobileuse} organizes reflection at action, trajectory, and task levels.
Digital-agent evaluators range from prompted step-level assessment~\citep{pan2024autonomous} to learned process reward models~\citep{chae2025webshepherd,chen2025guishepherd}.
\par
\fussy\tolerance=500\emergencystretch=0.5em\hyphenpenalty=20
GUI-Reflection~\citep{wu2025guireflection} trains verification and corrective behavior in end-to-end GUI agents; its Action Verification task checks a supplied action purpose against paired screenshots.
VisCritic~\citep{qian2026viscritic} predicts action success and task progress from visual state differences, while Task-State Representation~\citep{zheng2026taskstate} provides a training-free transition-aware wrapper.
EFR~\citep{ma2026evidencefirst} first extracts and filters observed visual changes before judging action success.
\tool formulates action reflection as supervised structured prediction conditioned on task context, action information, bounded history, and paired visual evidence.
It first generates action-specific pre- and post-conditions, then produces a structured rationale and a binary decision assessing execution success and task validity.
\par\endgroup
\par\endgroup

\section{\tool}
\label{sec:approach}
For GUI agents, a task $\tau$ specifies the user's goal. At step $t$, the agent executes action $a_t$ in UI state $s_t$ and reaches state $s_{t+1}$, forming a transition $(s_t,a_t,s_{t+1})$. Transition reflection evaluates whether the action achieves its intended local effect and whether the resulting transition constitutes a valid step toward $\tau$. Valid steps include preparation, navigation, information gathering, and recovery. Such steps need not complete the overall task.

\begingroup
\frenchspacing
\fussy\tolerance=1000\emergencystretch=1em\hyphenpenalty=30
\subsection{Problem Definition}
\begin{figure*}[t]
  \centering
  \includegraphics[width=\textwidth]{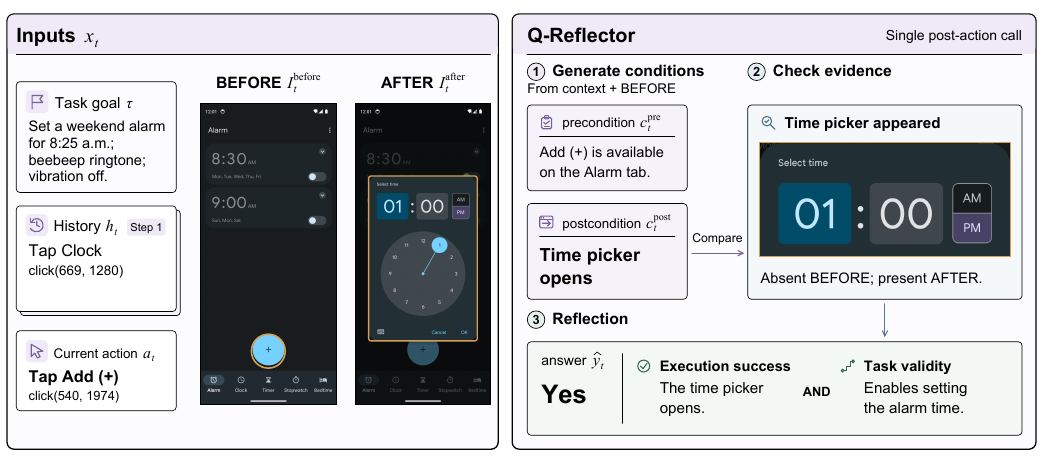}
  \caption{\textbf{Illustrative post-action reflection with \tool.}
  Task context, available history, the executed action, and paired
  BEFORE/AFTER screenshots form the input $x_t$. Within a single post-action
  call, \tool generates pre- and post-conditions $c_t^{\mathrm{pre}}$ and
  $c_t^{\mathrm{post}}$, then a rationale $r_t$ and a binary verdict $\hat{y}_t$.
  The rationale checks the conditions against visual evidence and assesses
  task validity. Opening the time picker is a valid preparatory step,
  although the alarm is not yet configured. Screenshots are from a recorded
  trajectory; the displayed conditions and explanation are illustrative.}
  \label{fig:qv}
\end{figure*}

\tool addresses the problem of  \emph{\textbf{GUI transition reflection}} for autonomous mobile GUI agents. Figure~\ref{fig:qv} illustrates the mapping from the transition input $x_t$ to generated pre- and post-conditions, a structured rationale, and a binary reflection decision.
After the GUI agent executes action $a_t$ in UI state $s_t$, \tool evaluates the resulting transition $s_t \rightarrow s_{t+1}$ to determine whether the action achieved its intended local effect and whether the transition constitutes a valid step toward the task goal $\tau$. Inspired by state-machine specifications of action transitions~\citep[Sec.~14.4.3.2]{omg2017uml}, we make an action's prerequisites and intended local effect explicit through generated pre- and post-conditions. These conditions provide action-specific criteria for checking the observed transition, while task validity is assessed against the task goal and recent history. We formulate GUI action reflection as supervised structured prediction over individual UI transitions. Conditioning the prediction on task context, action information, and paired visual evidence allows a compact VLM to learn action-level reflection independently of the action-generation policy.

Formally, at step $t$, the reflector receives the global task $\tau$, the executed action $a_t$, recent action history $h_t$, and the BEFORE/AFTER screenshot pair:
\[
x_t=\bigl(\tau,a_t,h_t,I_t^{\mathrm{before}},I_t^{\mathrm{after}}\bigr).
\]
The action $a_t$ comprises its type, arguments, and an intent description. This description expresses the intended operation rather than an observed state. The screenshots provide visual observations of $s_t$ and $s_{t+1}$, respectively. The task-conditioned reflector produces
\[
f_\theta(x_t)=\bigl(c_t^{\mathrm{pre}},c_t^{\mathrm{post}},r_t,\hat{y}_t\bigr),
\]
where $c_t^{\mathrm{pre}}$ and $c_t^{\mathrm{post}}$ are the generated pre- and post-conditions, respectively, $r_t\in\mathcal{R}$ is a rationale, and $\hat{y}_t\in\{0,1\}$ is the predicted verdict. These are serialized as the four fields \texttt{precondition}, \texttt{postcondition}, \texttt{reason}, and \texttt{answer}, in that order, with verdicts $1$ and $0$ represented by ``Yes'' and ``No'', respectively.
Let $E_t$ indicate that the intended local action effect is realized, and let $V_t$ indicate that the resulting transition is a valid task step given goal $\tau$ and history $h_t$. The target is $y_t=1$ if and only if $E_t\land V_t$; otherwise, $y_t=0$. A negative label can therefore indicate either unsuccessful execution or successful execution of a task-irrelevant action.
The reflection decision provides feedback for subsequent action generation; the execution loop determines whether to continue, retry, recover, or terminate execution.

\subsection{GUI Transition Representation}
\tool performs GUI action reflection at the granularity of \emph{\textbf{individual state transitions}}. Each reflection instance combines the global task, the executed action, bounded action history, and paired visual evidence.
\tool makes \emph{\textbf{semantic expectations}} explicit before checking \emph{\textbf{observed evidence}} in its response. It first generates action-specific pre- and post-conditions and then evaluates the observed transition. Both screenshots are supplied together; deriving conditions from the task, action, history, and BEFORE screenshot alone is an instruction, not a restriction on access to AFTER. The precondition specifies the prerequisites for the action, while the postcondition specifies its intended local effect.

The rationale then comprises three parts: \emph{Precondition check} assesses the starting conditions; \emph{Postcondition check} examines whether the in\-tend\-ed effect is supported by AFTER, using BEFORE to establish the change; and \emph{Task progress} assesses whether the observed effect constitutes a valid step toward the task goal. Action descriptions and history provide context whose claims must be checked against the available evidence. Generated preconditions are fallible hypotheses: an assumed prerequisite does not override clear evidence that the intended effect occurred. The model is instructed not to revise the expected postcondition to fit the observed outcome.

To incorporate \textit{\textbf{temporal context for reflection}}, \tool retains at most the five most recent actions:
\[
h_t=\mathrm{lastK}\!\left(\langle(i,a_i)\rangle_{i<t},\;h_{\max}\right),
\]
where entries are ordered by step index, $h_{\max}=5$, and $\mathrm{lastK}$ selects the most recent entries. Each entry contains only the step index and action; previous reflection outputs and historical screenshots are excluded. This bounded history provides local context for interpreting preparatory and recovery actions with respect to the task goal.

Finally, \tool is designed to
\textit{\textbf{integrate with existing GUI agents through a lightweight interface}}. Model-specific adapters map the available task context, action information, and visual observations to the structured representation defined in Section~3.1. When the parsed verdict is ``No'', the adapter appends a reflection hint to the action-generating model's next input. The hint identifies the previous action, includes the \emph{Postcondition check} and \emph{Task progress} portions of the rationale, and instructs the model to avoid repeating the same coordinates or text. Generated pre- and post-conditions are used within the reflector rather than directly passed to the action-generating model. If the verdict is ``Yes'' or cannot be parsed, no feedback is injected. For a ``No'' verdict, the adapter uses a shortened excerpt of the rationale if it cannot extract the \emph{Postcondition check} and \emph{Task progress} sections. The model uses this feedback for subsequent action generation.

\par\endgroup

\subsection{Learning a GUI Transition Reflector}
\begin{figure*}[t]
  \centering
  \includegraphics[width=\textwidth]{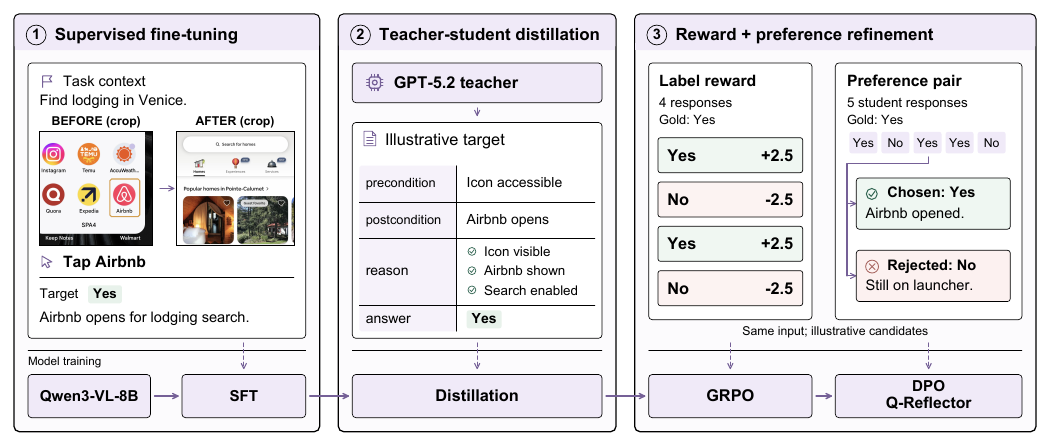}
  \caption{\textbf{Three-stage training of the final \tool model.}
  Supervised fine-tuning learns to generate pre-/post-conditions, structured rationales, and verdicts; teacher--student distillation continues fine-tuning on GPT-5.2 targets with loss over the complete four-field response. GRPO then applies asymmetric verdict rewards and a format bonus, followed by DPO on correct/incorrect response pairs sampled from the student for the same input. Screenshots show a training example; displayed responses are illustrative excerpts.}
  \label{fig:train}
\end{figure*}

\begingroup
\fussy\tolerance=1000\emergencystretch=1em\hyphenpenalty=30
\frenchspacing
We build \tool on Qwen3-VL-8B~\citep{bai2025qwen3vltechnicalreport}.
As shown in Figure~\ref{fig:train}, our fine-tuning pipe\-line comprises
three stages that progressively equip the model with transition
recognition and structured reflection capabilities.
\par\endgroup

\textbf{Stage 1: Supervised Fine-Tuning.}
We first apply supervised fine-tuning (SFT) to train \tool
to generate action-specific pre- and post-conditions
and assess execution success and task validity. Each training example instantiates the input $x_t$ defined above, and the model is trained to generate the complete four-field JSON response: precondition, postcondition, the three-part rationale, and final verdict. GPT-5.2 generates the pre- and post-condition targets, which are individually reviewed by human annotators. Rationale targets come from the collected trajectories and are also manually reviewed. The SFT loss covers all tokens in this target response. Before training, we verify for every example that the complete response is included in the supervised token span. This stage teaches the model to formulate action-specific expectations and relate them to visual evidence and task context when judging transitions, providing a stable optimization baseline.

\begingroup
\fussy\tolerance=1000\emergencystretch=1em\hyphenpenalty=30
\frenchspacing
\textbf{Stage 2: Teacher--Student Distillation.} We perform teacher--student distillation~\citep{ho2023reasoning} using GPT-5.2~\citep{openai_gpt52} as the teacher model. Each teacher response retains the four fields: precondition, postcondition, reason, and answer. We continue fine-tuning the SFT model on these responses. The distillation loss covers the complete four-field response, including both conditions, the rationale, and the final verdict. Distillation produces substantially longer, template-structured outputs, but it does not consistently preserve standalone classification accuracy.
\par\endgroup

\textbf{Stage 3: Preference and Reward-Based Refinement.}
To reduce incorrect rejection of valid transitions, we first apply Group Relative Policy Optimization (GRPO)~\citep{shao2024deepseekmath} with asymmetric verdict rewards and a format bonus. We then apply Direct Preference Optimization (DPO)~\citep{rafailov2024_dpo} to responses sampled from the student for the same input, using correct responses as chosen and incorrect responses as rejected. DPO optimizes the complete four-field responses against a frozen reference model; preferences depend on verdict correctness rather than independent condition or rationale quality scores. Appendix~\ref{app:training} details the rewards, pair construction, optimization settings, and computational resources.

\subsection{Training Data}
\label{sec:trainingData}
\begingroup\frenchspacing
\fussy\tolerance=1000\emergencystretch=1em\hyphenpenalty=30
We collect training trajectories by running Agent-SAMA~\citep{guo2025agentsama} on Mobile-Eval-E~\citep{wang2025mobileagente} and SPA-Bench~\citep{chen2025spabench}. The collected trajectories cover 45 English tasks across domains such as e-commerce, information management, and social media, spanning 40 apps. Each \emph{interaction step} in a trajectory is treated as an independent training instance.

At every step, we use the Reflection Agent design and prompting interface from Agent-SAMA~\citep{guo2025agentsama}, while replacing its original GPT-4o-2024-11-20 backbone with GPT-5.2. This GPT-5.2-based re-instantiation produces an initial judgment of whether the executed action results in the intended UI change, using the BEFORE/AFTER screenshots, UI state, task context, and action metadata. The authors then audit and correct these initial judgments to obtain author-audited transition-level reflection labels.

After filtering incomplete trajectories, the resulting training set contains 880 transition-level instances (512 positive, 58.18\%; 368 negative, 41.82\%). We evaluate offline reflection on AndroidWorld~\citep{rawles2024androidworld} and online task execution on MobileWorld~\citep{kong2025mobileworldbenchmarkingautonomousmobile}, both separate from the training benchmarks.
\par\endgroup

\section{Experiments}
\label{sec:eval}

We evaluate \tool through offline reflection on AndroidWorld
transitions and online integration with GUI agents on MobileWorld.

\subsection{Offline Reflection on AndroidWorld Trajectories}
\textbf{Setup.}
For offline evaluation, we use AndroidWorld~\citep{rawles2024androidworld},
which comprises 116 tasks involving one or more apps. Following the
process described in Section~\ref{sec:trainingData}, we construct a
test set of 1,082 manually verified transition-level instances,
including 613 positive samples (56.7\%) and 469 negative samples
(43.3\%). Each transition is evaluated independently according to whether the executed action produces its intended UI effect and the resulting transition constitutes a valid step toward the global task, given the recent action history. Both conditions are required for a positive label; a valid intermediate or preparatory step need not complete the task.

We report overall accuracy, positive-class accuracy for accepting valid transitions, negative-class accuracy for rejecting invalid transitions, and balanced accuracy, the unweighted mean of the two class accuracies. For the comparisons in Table~\ref{tab:offline_main}, the input consists of the global task, current action, up to five recent actions with their step indices, and BEFORE/AFTER screenshots. All reflectors receive the same transition inputs. All models use the same prompt, requesting \texttt{precondition}, \texttt{postcondition}, \texttt{reason}, and \texttt{answer}. Comparisons use the decoding settings in Appendix~\ref{app:inference}, with no additional API reasoning mode enabled.

The offline baselines include GPT-5.2, GPT-4o~\citep{openai2024gpt4o}, the Qwen variants~\citep{bai2025qwen3vltechnicalreport,bai2025qwen25vltechnicalreport}, MAI-UI-8B, GUI-Owl-1.5-8B, Claude Haiku 4.5~\citep{anthropic2025haiku45}, and GUI-Reflection-8B. MAI-UI-8B and GUI-Owl-1.5-8B serve as prompted reflectors. For GUI-Reflection, the comparison uses its Action Verification pretraining checkpoint with the same execution-success and task-validity criteria and its required image preprocessing. Reflection accuracy is computed from binary verdicts; explanations are not independently scored.

\begin{table}[t]
\centering
\small
\setlength{\tabcolsep}{2.5pt}
\begin{tabularx}{\columnwidth}{@{}>{\raggedright\arraybackslash}Xrrrr@{}}
\toprule
\textbf{Model} & \shortstack{\textbf{Overall}\\\textbf{Acc.}} & \shortstack{\textbf{Bal.}\\\textbf{Acc.}} & \shortstack{\textbf{Pos.}\\\textbf{Acc.}} & \shortstack{\textbf{Neg.}\\\textbf{Acc.}} \\
\midrule
Qwen2.5-VL-3B & 57.30 & 51.32 & 96.25 & 6.40 \\
Qwen2.5-VL-7B & 50.83 & 56.23 & 15.66 & 96.80 \\
Qwen3-VL-4B & 64.33 & 67.91 & 40.95 & 94.88 \\
Qwen3-VL-8B & 73.11 & 74.04 & 67.05 & 81.02 \\
\midrule
GPT-4o & 66.17 & 68.37 & 51.88 & 84.86 \\
GPT-5.2 & 70.33 & 73.24 & 51.39 & 95.10 \\
Claude Haiku 4.5 & 76.89 & 78.96 & 63.46 & 94.46 \\
MAI-UI-8B & 74.77 & 71.52 & 95.92 & 47.12 \\
GUI-Owl-1.5-8B & 77.36 & 75.06 & 92.33 & 57.78 \\
GUI-Reflection-8B & 60.72 & 57.02 & 84.83 & 29.21 \\
\midrule
\tool (Ours) & \textbf{82.16} & \textbf{81.45} & 86.79 & 76.12 \\
\bottomrule
\end{tabularx}
\caption{Offline baseline comparison on 1,082 AndroidWorld transitions (613 positive and 469 negative). All values are accuracies (\%). Qwen rows use base checkpoints; GUI-Reflection-8B uses its Action Verification pretraining checkpoint under the joint execution-success and task-validity criterion. Balanced accuracy weights the two classes equally. Bold marks the highest overall and balanced accuracies.}
\label{tab:offline_main}
\end{table}

\textbf{Offline reflection performance.}
With condition generation, \tool achieves 82.16\% overall accuracy,
exceeding all baselines in Table~\ref{tab:offline_main}. The strongest
baseline, GUI-Owl-1.5-8B, reaches 77.36\%, while GPT-5.2 achieves
70.33\%. Balanced accuracy reaches 81.45\%, compared with 78.96\%
for Claude Haiku 4.5, the strongest baseline on this metric.

\subsection{Online End-to-End Evaluation with Integrated Reflection}

\textbf{Setup.}
We conduct online evaluation on MobileWorld~\cite{kong2025mobileworldbenchmarkingautonomousmobile} using MAI-UI-8B~\cite{zhou2025maiuitechnicalreportrealworld} and GUI-Owl-1.5-8B-Instruct~\cite{xu2026mobileagentv35} as action-generating models.

For MobileWorld, all settings use the same 117 tasks. We compare no reflection, self-reflection, and reflection with GPT-5.2, Claude Haiku 4.5, GUI-Reflection-8B, and \tool. No reflection adds no separate reflection call; it does not exclude reflection within the model's action-generation process. Self-reflection prompts the same checkpoint in a separate call, using the same input and feedback protocol as the other reflectors. These comparisons assess the effect of adding explicit reflection to task execution. Each reflector receives the global task, current action, up to five recent actions with their step indices, and BEFORE/AFTER screenshots, and judges both execution success and task validity. For each action-generating model, the comparison holds the input information and feedback interface fixed: a negative verdict triggers a reflection hint derived from the explanation, which is appended to the next observation as described in Section~\ref{sec:approach}.
Table~\ref{tab:online_results} groups results by action-generating model and reflection setting. Without condition generation, the reflector produces a rationale and binary verdict. With condition generation, it first generates action-specific pre-/post-conditions, followed by the rationale and verdict. For \tool, the two settings use separately trained variants with matched training configurations (Section~\ref{sec:condition_ablation}); condition generation is part of the full method.

\textbf{End-to-end performance.}
With GUI-Owl-1.5-8B-Instruct generating actions, \tool increases task
success from 29.06\% without external reflection to 46.15\%.
With MAI-UI-8B, task success increases from 21.37\% to 35.90\%.
These improvements correspond to 20 and 17 additional completed tasks,
respectively. Under the same condition-generation setting, \tool also
achieves higher observed success than the strongest baseline reflector for each action model:
GPT-5.2 at 42.74\% with GUI-Owl-1.5-8B-Instruct and Claude Haiku 4.5
at 29.91\% with MAI-UI-8B.

A qualitative example in Appendix~\ref{app:brightness-case}
(Figure~\ref{fig:brightness-case}) contrasts repeated ineffective
swipes without reflection with recovery at the next step following
feedback from \tool.

\textbf{Token usage per task.} Table~\ref{tab:online_results} presents task success alongside mean input and output tokens per attempted task. Token usage includes action-generation and reflection calls across all 117 tasks, including failed tasks, recovery calls, and retries. Both locally served and hosted reflectors contribute to these totals. Local inference uses the model's tokenizer or serving-backend counts, while hosted inference uses API usage fields. Visual and cached input tokens contribute to the input total, and provider-reported reasoning tokens contribute to the output total where applicable, without double counting. Visible explanations are part of generated output. These counts describe token usage under each backend's tokenizer and visual encoder, rather than a standardized computational cost.

\begin{table*}[t]
\centering
\small
\setlength{\tabcolsep}{4pt}
\begin{tabularx}{\textwidth}{@{}>{\raggedright\arraybackslash}p{0.24\textwidth}*{6}{>{\raggedleft\arraybackslash}X}@{}}
\toprule
& \multicolumn{3}{c}{\textbf{Without condition generation}}
& \multicolumn{3}{c}{\textbf{With condition generation}} \\
\cmidrule(lr){2-4}\cmidrule(lr){5-7}
\textbf{Reflector}
& \textbf{SR (\%)} & \shortstack{\textbf{Input}\\\textbf{tok./task}} & \shortstack{\textbf{Output}\\\textbf{tok./task}}
& \textbf{SR (\%)} & \shortstack{\textbf{Input}\\\textbf{tok./task}} & \shortstack{\textbf{Output}\\\textbf{tok./task}} \\
\midrule
\multicolumn{7}{@{}l}{\textit{Action model: MAI-UI-8B; MobileWorld, 117 tasks}} \\
No reflection & 21.37 & 335,085 & 2,239 & - & - & - \\
Self-reflection & 30.77 & 361,564 & 4,083 & 27.35 & 448,438 & 9,951 \\
GPT-5.2 & 24.79 & 393,555 & 3,923 & 29.06 & 465,816 & 10,561 \\
Claude Haiku 4.5 & 31.62 & 387,943 & 7,151 & 29.91 & 456,439 & 16,808 \\
GUI-Reflection-8B & 24.79 & 342,591 & 2,003 & 29.06 & 445,131 & 5,691 \\
\tool & 29.91 & 261,933 & 6,572 & \textbf{35.90} & 460,894 & 9,912 \\
\midrule
\multicolumn{7}{@{}l}{\textit{Action model: GUI-Owl-1.5-8B-Instruct; MobileWorld, 117 tasks}} \\
No reflection & 29.06 & 143,649 & 1,907 & - & - & - \\
Self-reflection & 39.32 & 166,453 & 3,141 & 41.88 & 229,996 & 8,031 \\
GPT-5.2 & 40.17 & 174,822 & 3,472 & 42.74 & 234,310 & 8,924 \\
Claude Haiku 4.5 & 35.90 & 204,269 & 7,755 & 39.32 & 240,750 & 14,858 \\
GUI-Reflection-8B & 34.19 & 184,933 & 2,128 & 39.32 & 242,078 & 4,988 \\
\tool & 41.88 & 167,299 & 8,562 & \textbf{46.15} & 236,132 & 8,082 \\
\bottomrule
\end{tabularx}
\caption{Online task success (SR) and mean token usage on MobileWorld (117 tasks per action model). Condition generation denotes reflector-generated pre-/post-conditions preceding evidence checks and the verdict. Q-Reflector uses separately trained variants with matched training configurations; other reflectors use fixed checkpoints. Tokens include action-generation and reflection calls across all tasks, including failures and retries. No reflection adds no separate reflection call; self-reflection uses the same checkpoint in a separate call. A dash denotes an inapplicable measurement; condition generation is not applicable to the no-reflection baseline.}
\label{tab:online_results}
\end{table*}

\subsection{Ablation of Condition Generation}
\label{sec:condition_ablation}

\textbf{Setup.}
We ablate pre- and post-condition generation in the online evaluation
(Table~\ref{tab:online_results}). The full \tool procedure generates
these conditions before the rationale and verdict; the ablation produces
only the rationale and verdict. The variants are trained separately from
the same base model using the same 880 training transitions, training
stages, and hyperparameters. Only the prompts and supervised response
structures differ. Evaluation inputs and the feedback interface are held fixed.

\textbf{Effect on task success.}
With GUI-Owl-1.5-8B-Instruct generating actions, the full \tool achieves
46.15\% task success, exceeding the ablation's 41.88\% by
4.27 percentage points. With MAI-UI-8B, condition generation increases
task success from 29.91\% to 35.90\%, corresponding to seven additional
completed tasks. The condition-generating variant achieves higher task success with both
action models. For the other reflectors, whose checkpoints remain fixed,
prompting condition generation also improves success with GUI-Owl-1.5-8B-Instruct,
completing three to six additional tasks. With MAI-UI-8B, the other
reflectors show mixed effects:
GPT-5.2 and GUI-Reflection-8B each complete five additional tasks,
while self-reflection and Claude Haiku 4.5 complete four and two fewer
tasks, respectively. The benefit of condition generation thus depends
on the combination of action model and reflector.

\textbf{Effect on token usage.}
Compared with the ablation without condition generation, the full \tool
increases mean input token usage per task by 41.1\% with
GUI-Owl-1.5-8B-Instruct and 76.0\% with MAI-UI-8B.
Output usage decreases by 5.6\% with GUI-Owl-1.5-8B-Instruct but
increases by 50.8\% with MAI-UI-8B.
The expanded prompt introduces instructions for deriving pre- and
post-conditions, and the response includes these additional fields before
the rationale and verdict. Per-task token usage also depends on the
number of executed actions and reflection calls, so the observed changes
reflect both the additional condition-generation requirements and the
resulting execution trajectories.

\section{Limitations}
\label{sec:limitation}

\textbf{Data and label scalability.}
GPT-5.2-assisted reflection labels are manually audited and corrected, and SFT pre- and post-condition targets are individually reviewed. This human effort limits scalability. Evaluation measures verdict accuracy and downstream task success, without independently scoring the semantic correctness of generated conditions.

\textbf{Evaluation and deployment scope.}
We evaluate 1,082 AndroidWorld transitions and 117 MobileWorld tasks. Both online action-generating models, MAI-UI-8B and GUI-Owl-1.5-8B-Instruct, have 8B parameters; transfer to other sizes remains untested. Per-task input and output token totals include local and API-based action generation and reflection. Counts depend on tokenizers and visual encoders and do not directly measure latency or hardware cost. Latency also depends on serving hardware, batching, and API overhead. Limited history may obscure the purpose of preparatory or recovery steps when assessing task validity.

\textbf{Cross-platform transfer.}
Validation focuses on mobile GUI agents. Web and desktop workflows can involve browser tabs, multiple windows, and keyboard shortcuts, for which the appropriate action-specific pre- and post-conditions may differ from those in our mobile tasks. Whether the trained reflector handles these interactions reliably or condition generation yields comparable gains remains untested.

\textbf{Screenshot preprocessing.}
Online reflection uses aspect-ratio-preserving screenshot resizing with a maximum side length of 1,024 pixels, followed by model-specific image processing. The reported results reflect this preprocessing protocol. We do not systematically evaluate alternative resizing limits or their effects on reflection accuracy and downstream task success.

\textbf{Decoding configuration.}
Online evaluation uses fixed decoding settings for each model and a 1,024-token limit for reflection outputs. The same output limit applies with and without condition generation. We do not systematically evaluate alternative sampling temperatures or output budgets, so sensitivity to these inference choices remains untested.

\section{Conclusion}
\label{sec:conclusion}

We presented \tool, a GUI reflector based on Qwen3-VL-8B, trained to
assess execution \mbox{success} and task validity. It generates action-specific
pre- and post-conditions before checking the observed transition and
producing a rationale and binary verdict. On AndroidWorld transitions,
\tool achieves 82.16\% accuracy with more balanced recognition of valid
and invalid transitions than GPT-5.2 and GUI-Owl-1.5-8B. On MobileWorld,
\tool achieves task success rates of 46.15\% with GUI-Owl-1.5-8B-Instruct
and 35.90\% with MAI-UI-8B. For both action-generating models, \tool
improves over no external reflection and achieves the highest observed
success among reflectors evaluated under the common condition-generation
procedure.
\tool combines accurate, balanced judgments learned through
task-specific training with explicit conditions that further improve
task execution.

\section{Ethical Considerations}
\label{sec:ethics}

\tool assesses action execution and task validity, but its verdicts do not establish user authorization or operational safety. Incorrect reflection may lead to unintended GUI actions; consequential operations, such as payments or data deletion, should require explicit user confirmation. Screenshots and action histories may contain sensitive information, especially when sent to external model services. Deployment should minimize data collection and retention and obtain appropriate consent for transmission. Our benchmark evaluation does not establish safety or privacy guarantees for real-world deployment.

\bibliography{custom}

\onecolumn
\raggedbottom
\appendix

\frenchspacing
\fussy\tolerance=1500\emergencystretch=1em\hyphenpenalty=30

\lstdefinestyle{prompt}{
  basicstyle=\ttfamily\scriptsize,
  columns=fullflexible,
  keepspaces=true,
  breaklines=true,
  breakindent=0pt,
  frame=none,
  language={},
  showstringspaces=false,
  aboveskip=2pt,
  belowskip=2pt
}

\tcbset{appendixbox/.style={colback=gray!3,colframe=black!70,
  colbacktitle=gray!10,coltitle=black,boxrule=0.5pt,arc=2pt,
  title={#1},fonttitle=\bfseries\small,fontupper=\small}}
\newtcolorbox{appendixpromptbox}[1]{appendixbox={#1},breakable}
\newtcolorbox{widepromptbox}[1]{appendixbox={#1}}
\lstdefinestyle{wideprompt}{style=prompt,
  basicstyle=\ttfamily\scriptsize,breakatwhitespace=true}

\section{Data Construction and Annotation}
\label{app:data}

\subsection{Training and evaluation separation}
We run Agent-SAMA on Mobile-Eval-E and SPA-Bench, covering 45 English
cross-domain tasks and 40 apps, and collect actual execution transitions.
After filtering and human audit, the training set contains 880 transitions:
512 positive (58.18\%) and 368 negative (41.82\%). A separate validation set
contains 71 transitions from the same sources, held out at the transition
level without overlap with the training set.
Offline evaluation uses the separate AndroidWorld benchmark. Its frozen panel
contains 1,082 manually verified transitions: 613 positive (56.7\%) and 469
negative (43.3\%). No AndroidWorld transition is used for SFT, teacher
generation, DPO-pair construction, or GRPO updates. Online evaluation uses
117 MobileWorld tasks, also separate from the training benchmarks.
No evaluation outcome is used to construct training examples or select the
evaluation set. Held-out labels, rewards, and task results are excluded from
training and deployment prompts.

\subsection{Label audit}
The judgments and rationales collected from Agent-SAMA are reviewed against
paired screenshots and the action's intended local effect. Human audit
corrects cases that confuse intent, intermediate progress, visual evidence, or
task completion. Incomplete records are filtered rather than guessed. The audit
verifies image--step alignment, identifies visible changes independently of
the draft, checks the intended local effect, and requires rationales grounded
in observable evidence.

\section{Training Details}
\label{app:training}

\subsection{Optimization and supervision}
SFT for \tool uses BF16 model weights
and LoRA with $r=16$, $\alpha=16$, zero dropout, no bias, and seed 42. Adapters
target \texttt{q\_proj}, \texttt{k\_proj}, \texttt{v\_proj}, \texttt{o\_proj},
\texttt{gate\_proj}, \texttt{up\_proj}, and \texttt{down\_proj}. Gradient
checkpointing is enabled. Table~\ref{tab:app-train-config} summarizes the
per-stage configuration.

\begin{table}[H]
\centering
\small
\setlength{\tabcolsep}{3pt}
\renewcommand{\arraystretch}{1.15}
\begin{tabularx}{\columnwidth}{@{}lccc>{\raggedright\arraybackslash}X@{}}
\toprule
\textbf{Stage} & \textbf{Epochs} & \textbf{LR} & \shortstack{\textbf{Eff.}\\\textbf{batch}} & \textbf{Key setting} \\
\midrule
SFT     & 15 & $2\times10^{-4}$ & 8  & cosine, 0.1 warmup \\
TS-Dis. & 3 & $1\times10^{-4}$ & 8  & GPT-5.2 targets \\
GRPO    & 1 & $1\times10^{-6}$ & 4  & 4 generations \\
DPO     & 3 & $5\times10^{-6}$ & 16 & $\beta=0.1$ \\
\bottomrule
\end{tabularx}
\caption{Training configuration for each stage.}
\label{tab:app-train-config}
\end{table}

We select the epoch-9 SFT checkpoint based on validation accuracy.
Training loss is used to monitor training.
SFT and distillation use 8-bit AdamW, bfloat16 computation, cosine scheduling,
and 0.1 warmup. GPT-5.2 generates the SFT condition targets, each reviewed
by human annotators. SFT rationales come from the collected reasoning and
are also manually reviewed. Distillation continues from SFT using GPT-5.2
four-field responses; conflicts between teacher verdicts and audited labels
are manually checked. Both stages supervise all response tokens: conditions,
the three-part rationale, and the verdict. Before training, we verify that
each serialized target lies fully within the supervised span.

The condition-generation ablation is trained separately with the same base
model, 880 training transitions, inputs, training stages, and hyperparameters.
Only the prompt and supervised response structure differ: the ablation uses
\texttt{reason} and \texttt{answer} throughout training and inference.
The objectives below describe the full four-field model.

\paragraph{Computational resources.}
Each recorded local run uses one NVIDIA H200 GPU. The four stages of the
reported full \tool training run total 49.91 GPU-hours; five recorded
offline evaluations total 17.23 GPU-hours.
Seven MobileWorld runs take 5.44--7.34 GPU-hours each (45.63 in total).
These estimates use elapsed runtime.

\subsection{GRPO}
Audited verdict labels are used for reward computation and are not part of
the model input. Four completions are drawn per prompt at temperature 0.8. The base reward is
\begin{equation}
R(\hat{y}, y) =
\begin{cases}
+2.5, & \hat{y}=\text{Yes},\ y=\text{Yes},\\
+2.0, & \hat{y}=\text{No},\ y=\text{No},\\
-2.0, & \hat{y}=\text{Yes},\ y=\text{No},\\
-2.5, & \hat{y}=\text{No},\ y=\text{Yes}.
\end{cases}
\end{equation}
The stronger false-negative penalty reflects the cost of unnecessary recovery or
termination. A 0.1 format bonus requires nonempty \texttt{precondition} and
\texttt{postcondition}, the specified four-field order, and the three-part
\texttt{reason}; otherwise, it is zero. The ablation uses the same verdict
rewards and bonus, checking its two-field \texttt{reason}--\texttt{answer}
format. The reward does not assess condition quality, including
consistency with BEFORE or the action, or independence from AFTER.
The GRPO objective is computed over all generated response tokens,
including the precondition, postcondition, rationale, and final answer.
Prompt and padding tokens are excluded from the loss.

\subsection{Rejection-sampled DPO}
For each input and BEFORE/AFTER image pair, we sample five responses from
the student model at temperature 0.8. Responses whose final verdicts match
the audited label form the chosen pool; those with the opposite verdict
form the rejected pool. Both must satisfy the output format and be
untruncated. After deduplication, we retain at most two eligible rejected
responses in generation order and pair each with a correct response sampled
for the same input; rejected responses are not ranked by difficulty.
Inputs with unresolved label disagreements or without eligible responses
in both pools are excluded.
Applying this procedure to the 880 training transitions yields 559 valid
preference pairs. The retained pairs are checked for the complete four-field
structure, correct chosen/rejected label assignment, and matching image
references for the input transition.

Both responses retain all four fields. Preference pairs are selected by
final-verdict correctness, without independent semantic-quality scores for
conditions or rationales; correct verdicts do not guarantee correct explanations.

We apply standard DPO with a frozen reference model. For both the policy and
reference model, sequence log-probabilities are summed over all response
tokens, including the first response token and EOS, while excluding prompt
and padding tokens. Hyperparameters are listed in
Table~\ref{tab:app-train-config}.

\section{Prompts and Feedback Interface}
\label{app:prompts}
\makeatletter
\setlength{\@fptop}{0pt}
\makeatother

\subsection{Input serialization}
Both settings use the \texttt{\{context\}} object shown below.
History contains up to five chronological step/action records; all available
records are retained if there are fewer than five, and the initial history is
empty. No separate screen descriptions, subgoals, or condition fields are supplied.

\begin{appendixpromptbox}{Input context: runtime substitution}
\begin{lstlisting}[style=prompt]
{
  "goal": "<complete task goal>",
  "action": {
    "type": "<action type>",
    "arguments": <action arguments>,
    "description": "<action intent>"
  },
  "history": <ordered step/action records>
}
\end{lstlisting}
\end{appendixpromptbox}

Both settings receive identical geometry metadata: the original and resized
image dimensions and, when applicable, normalized coordinates for the current
action. This metadata locates the action target after resizing; it is not
evidence of execution success.

\subsection{Reflection prompts}
Both screenshots are supplied in the same call. Forming conditions from
context and BEFORE alone is an instruction to the model, not a runtime
separation of its access to the two images.

\textbf{With condition generation.}
The prompt requests pre- and post-conditions, a rationale with three labeled
checks, and a binary verdict, in that order
(Figure~\ref{fig:app-prompt-with}).

\begin{figure}[p]
\centering
\begin{widepromptbox}{With condition generation}
\begin{lstlisting}[style=wideprompt]
You are a mobile GUI reflector performing post-action verification. Judge whether this single action was correctly executed AND made valid progress toward the user's task. Both requirements must be satisfied for "Yes". Useful intermediate progress counts; completing the entire task is not required. Action descriptions and history may contain mistakes. Use the task, the actual operation, and visible evidence rather than accepting those claims as facts.

AFTER is the primary evidence of the action's outcome. BEFORE establishes the starting state and helps identify what changed, which target was affected, and whether the apparent result was already present. Use both images to assess the transition; a desirable AFTER state or an arbitrary screen change alone is not proof that this action succeeded or advanced the task.

Follow this procedure in the required output order:

Define both conditions using only the goal, current action, available history, and BEFORE screenshot. They are verification criteria, not descriptions of AFTER. For the same starting state and action, the expected postcondition must remain the same whether AFTER shows success or failure. Use AFTER only in the checks and verdict below, never to construct or retrospectively change the expectation.

1. "precondition": Specify the action-specific conditions that need to hold in the BEFORE state for the intended action to be carried out. Derive them from the goal, action, history, and BEFORE, without using AFTER. These are conditions on the starting state, not predictions of the resulting state or a general screenshot description. Respect the action type: Open_App does not require a visible launcher icon. Do not invent prerequisites or treat a prerequisite not visible in BEFORE as necessarily absent.

2. "postcondition": Specify what should become true or change if THIS action is correctly executed from the BEFORE state. Derive this expected local result from the same context and BEFORE, without using AFTER. Match the action type, intended target, and supplied value. Use the goal to establish relevance, not to demand later steps. For example, opening an app should bring it to the foreground; a long press intended to select text should activate selection or text-editing controls, not already replace the text or save the document. Define the expectation independently of AFTER; do not rewrite it to fit an observed failure.

3. "reason": Use these three brief labeled parts:
- Precondition check: Check the stated starting conditions against the context and BEFORE; distinguish contradicted from unobservable. Your generated prerequisites are fallible hypotheses, not additional success requirements. If AFTER clearly demonstrates the intended effect, explain in this check whether a conflicting assumed prerequisite was mistaken; do not let that assumption automatically veto success or rewrite the expected postcondition.
- Postcondition check: Locate evidence for the expected effect in AFTER, using BEFORE to establish the change. Check the actual app, control, field, and text or numeric value as relevant. State whether the expected effect is supported, contradicted, or unobservable, with specific evidence.
- Task progress: Judge whether that observed effect advances the task, including a useful preparatory step. Do not demand work belonging to later actions or excuse an incorrect target or value as partial progress.

4. "answer": Give "Yes" only when the evidence supports both correct execution of the intended local effect and useful task progress; otherwise give "No". Base the verdict on the observed outcome and the checks, not on whether you produced conditions. Keep the verdict and the described evidence consistent.

Both screenshots are provided together in this single post-action call.
=== Image Geometry ===
The images may be resized. Current-action coordinates are pixels in the original BEFORE screenshot, with origin at top-left, x increasing rightward and y increasing downward. Normalized coordinates are (x / original_width, y / original_height): (0, 0) is top-left and (1, 1) is bottom-right. Use the normalized positions to locate action targets regardless of resizing. Do not interpret original-image pixel coordinates as resized-image pixels. This is geometric metadata, not evidence of success or the identity of a control.
{image_geometry}
=== End Image Geometry ===
=== Context ===
{context}
=== Image Reference ===
<image_0>: Image BEFORE the action
<image_1>: Image AFTER the action
=== Required Output ===
Return only one JSON object with these string fields in this order:
"precondition": conditions required before this action, derived from context and BEFORE only;
"postcondition": the expected result of correctly executing this action from BEFORE, defined without using AFTER;
"reason": "Precondition check: ... Postcondition check: ... Task progress: ...";
"answer": exactly "Yes" or "No".
\end{lstlisting}
\end{widepromptbox}
\caption{Complete reflection prompt with condition generation.}
\label{fig:app-prompt-with}
\end{figure}

\textbf{Without condition generation.}
The prompt requests only \texttt{reason} and \texttt{answer}, without
explicit condition generation or the three labeled checks. The input
information and feedback interface remain unchanged
(Figure~\ref{fig:app-prompt-without}). For \tool, this prompt is used to train
and evaluate the separate ablation model; other reflectors use their existing
checkpoints under each prompt.

\begin{figure}[t]
\centering
\begin{widepromptbox}{Without condition generation}
\begin{lstlisting}[style=wideprompt]
You are a mobile GUI reflector performing post-action verification. Judge whether this single action was correctly executed AND made valid progress toward the user's task. Both requirements must be satisfied for "Yes". Useful intermediate progress counts; completing the entire task is not required. Action descriptions and history may contain mistakes. Use the task, the actual operation, and visible evidence rather than accepting those claims as facts.

AFTER is the primary evidence of the action's outcome. BEFORE establishes the starting state and helps identify what changed, which target was affected, and whether the apparent result was already present. Use both images to assess the transition; a desirable AFTER state or an arbitrary screen change alone is not proof that this action succeeded or advanced the task.

Follow this procedure in the required output order:

1. "reason": Give a brief evidence-based rationale for whether the action was correctly executed and made useful task progress. Locate evidence of the action's local effect in AFTER, using BEFORE to establish what changed. Check the actual app, control, field, and text or numeric value as relevant. Distinguish what the evidence supports, contradicts, or leaves unobservable. Respect the action type: Open_App does not require a visible launcher icon. Do not invent requirements or treat something not visible in BEFORE as necessarily absent. An assumption about BEFORE must not automatically veto an intended effect clearly demonstrated in AFTER. Use the goal to establish relevance, not to demand later steps. For example, opening an app should bring it to the foreground; a long press intended to select text should activate selection or text-editing controls, not already replace the text or save the document. Useful preparatory steps count as progress, but do not excuse an incorrect target or value as partial progress.

2. "answer": Give "Yes" only when the evidence supports both correct execution of the intended local effect and useful task progress; otherwise give "No". Base the verdict on the observed outcome and the rationale. Keep the verdict and the described evidence consistent.

Both screenshots are provided together in this single post-action call.
=== Image Geometry ===
The images may be resized. Current-action coordinates are pixels in the original BEFORE screenshot, with origin at top-left, x increasing rightward and y increasing downward. Normalized coordinates are (x / original_width, y / original_height): (0, 0) is top-left and (1, 1) is bottom-right. Use the normalized positions to locate action targets regardless of resizing. Do not interpret original-image pixel coordinates as resized-image pixels. This is geometric metadata, not evidence of success or the identity of a control.
{image_geometry}
=== End Image Geometry ===
=== Context ===
{context}
=== Image Reference ===
<image_0>: Image BEFORE the action
<image_1>: Image AFTER the action
=== Required Output ===
Return only one JSON object with exactly these two string fields in this order:
"reason": a brief evidence-based rationale about execution correctness and task progress;
"answer": exactly "Yes" or "No".
\end{lstlisting}
\end{widepromptbox}
\caption{Complete reflection prompt without condition generation, used in the online ablation.}
\label{fig:app-prompt-without}
\end{figure}

\subsection{Feedback interface}
\label{app:online}
Only a parsed \texttt{No} verdict triggers the following hint. The previous
action is combined with the \emph{Postcondition check} and \emph{Task progress}
sections extracted from \texttt{reason}. A \texttt{Yes} or unparseable verdict
produces no feedback.

The standalone condition fields and \emph{Precondition check} are not directly
forwarded; \texttt{answer} controls the trigger rather than appearing in the
hint. If the labeled checks cannot be extracted, the adapter uses a truncated
explanation. The same fallback handles an unsegmented \texttt{reason} in the
ablation, retaining the previous action and instruction to avoid repetition.
The execution loop controls subsequent actions, recovery, and termination.
The adapter leaves policy weights, grounding, action space, emulator, and
evaluator unchanged.

\begin{appendixpromptbox}{Feedback after a negative verdict}
\begin{lstlisting}[style=prompt]
<reflection_hint attempt="1">
Previous action: <executed action>.
Postcondition check: <outcome judgment and evidence>.
Task progress: <task-relevance judgment>.
Avoid repeating the same coordinates or text.
</reflection_hint>
\end{lstlisting}
\end{appendixpromptbox}

\clearpage
\section{Supplementary Experiments}
\label{app:experiments}
\makeatletter
\setlength{\@fptop}{0pt}
\makeatother

\subsection{Inference and online execution}
\label{app:inference}

Table~\ref{tab:app-inference-models} lists online checkpoints and decoding settings.
GUI-Owl-1.5-8B abbreviates GUI-Owl-1.5-8B-Instruct.
Self-reflection calls the same checkpoint as action generation.
These settings apply with and without condition generation
(Table~\ref{tab:online_results}).

\begin{table}[htbp]
\centering
\small
\setlength{\tabcolsep}{4pt}
\renewcommand{\arraystretch}{1.12}
\begin{tabularx}{\textwidth}{@{}>{\raggedright\arraybackslash}p{0.17\textwidth}>{\raggedright\arraybackslash}X>{\raggedright\arraybackslash}p{0.29\textwidth}@{}}
\toprule
\textbf{Model} & \textbf{Checkpoint or API model ID} & \textbf{Decoding} \\
\midrule
MAI-UI-8B & \texttt{Tongyi-MAI/\allowbreak MAI-UI-8B}; revision \texttt{e00a0097abb9} & Temperature 0, \texttt{top\_p}=1 \\
GUI-Owl-1.5-8B & \texttt{mPLUG/\allowbreak GUI-Owl-1.5-8B-Instruct}; revision \texttt{06d5faecff74} & Temperature 0, \texttt{top\_p}=1 \\
GPT-5.2 & \texttt{gpt-5.2-2025-12-11} & Temperature 0, \texttt{top\_p}=1; \texttt{reasoning\_effort=none} \\
Claude Haiku 4.5 & \texttt{claude-haiku-4-5-20251001} & Temperature 0; extended thinking disabled \\
GUI-Reflection-8B & \texttt{craigwu/\allowbreak GUI\_Reflection\_8b\_pretrain}; revision \texttt{04c75c2e66c0} & Greedy: \texttt{do\_sample=False}, \texttt{num\_beams=1} \\
\tool & Qwen3-VL-8B-Instruct with rank-16 LoRA & Greedy: \texttt{do\_sample=False}, \texttt{num\_beams=1} \\
\bottomrule
\end{tabularx}
\caption{Model versions and decoding settings for online evaluation.
Revisions are unique commit prefixes. Local models use BF16 inference;
GUI-Reflection uses its Action Verification pretraining checkpoint.
MAI and GUI-Owl use the stated decoding settings for both action generation
and self-reflection.}
\label{tab:app-inference-models}
\end{table}

Action generation is limited to 2,048 output tokens and reflection to 1,024.
Offline reflection uses the same decoding settings and reflection output limit;
GPT-4o uses \texttt{gpt-4o}, temperature~0, and \texttt{top\_p}=1.

\paragraph{Image processing.}
Action-generating models receive PNG-encoded original screenshots, typically
$1080\times2400$, without client-side resizing. Each model then applies its
own image processor.
For every reflector, BEFORE and AFTER are converted to RGB and resized
independently with Lanczos interpolation, preserving aspect ratio and
limiting the longest side to 1,024 pixels
($1080\times2400\rightarrow460\times1024$).
Qwen-based models then use their respective processors. GUI-Reflection uses
the official InternVL dynamic tiling with \texttt{image\_size}=448,
\texttt{max\_num}=6, and a thumbnail. Cloud providers control any subsequent
internal image processing.

\paragraph{Execution and scoring.}
Each task attempt permits at most 50 action-model decision steps, including
the final termination or answer action; reflector calls do not consume
additional decision steps. An attempt ends on a termination or answer action, the step
limit, or an unrecoverable parsing or execution error. Consecutive
\texttt{No} verdicts do not terminate an attempt
(\texttt{reflection\_\allowbreak max\_\allowbreak consecutive\_\allowbreak failures}=0).
After normal termination, MobileWorld's evaluator determines task success;
a model's completion claim alone is insufficient.
Each configuration covers all 117 tasks and retains one final score per task.
Table~\ref{tab:app-retries} specifies retries. Negative verdicts use the
feedback interface in Appendix~\ref{app:online}; the action-generating model
selects the next action within the remaining step budget.

\begin{table}[htbp]
\centering
\small
\setlength{\tabcolsep}{4pt}
\renewcommand{\arraystretch}{1.12}
\begin{tabularx}{\textwidth}{@{}>{\raggedright\arraybackslash}p{0.23\textwidth}>{\raggedright\arraybackslash}X@{}}
\toprule
\textbf{Failure or retry level} & \textbf{Handling} \\
\midrule
Action-model API request & 120-s timeout; at most three total request attempts, with 1-s intervals after failure. SDK retries are disabled. \\
Action-output parsing & GUI-Owl permits up to five generation/parse attempts. MAI returns \texttt{UNKNOWN} and ends the trajectory on a parsing failure. \\
API-based reflection & One request per verification. GPT-5.2 and local OpenAI-compatible self-reflection use 60-s timeouts; Haiku uses 120~s. API errors or missing valid verdicts are recorded as unavailable, produce no negative feedback, and allow execution to continue. \\
Local reflector worker & \tool and GUI-Reflection have a 300-s request wait limit. Worker exceptions propagate to task-level exception handling and do not produce a verdict. \\
Emulator or broker & Specified transient failures, such as unhealthy devices or HTTP~500 responses, trigger task reinitialization after 30~s, with at most five retries. Each initialization restarts the experiment container. \\
Missing task result & The outer scheduler permits at most ten total rounds (\texttt{auto\_retry}=10), rescheduling only tasks without a result. Tasks with a normal failure score are not automatically rerun. \\
\bottomrule
\end{tabularx}
\caption{Request, execution, and scheduling retries used in both online
condition-generation settings. Reflection feedback prompts a new action
choice; it does not automatically replay the previous action.}
\label{tab:app-retries}
\end{table}

\subsection{Evaluation metrics}
\label{app:offline}

For the offline test set, let $P=613$ and $N=469$. The metrics are
\begin{gather}
\mathrm{Acc}=\frac{TP+TN}{P+N},\\
\mathrm{PosAcc}=\frac{TP}{P},\quad
\mathrm{NegAcc}=\frac{TN}{N},\\
\mathrm{BalancedAcc}=\frac{\mathrm{PosAcc}+\mathrm{NegAcc}}{2}.
\end{gather}
Overall accuracy weights classes by their sample counts, whereas balanced
accuracy weights the two classes equally. Both are computed from binary
verdicts; rationales are not independently scored.
Online token usage follows the accounting in Section~\ref{sec:eval}; separate
action-generation and reflection counts are unavailable for the
condition-generation runs.

\subsection{Backbone and training-stage comparison}

Table~\ref{tab:app-backbone} compares base, SFT, and distilled checkpoints.
SFT improves overall accuracy for every backbone, while the effect of
subsequent distillation varies. For Qwen3-VL-8B, overall accuracy rises from
73.11\% to 80.87\% after SFT and decreases to 78.37\% after distillation.

\begin{table}[htbp]
\centering
\small
\setlength{\tabcolsep}{3pt}
\begin{tabular*}{\textwidth}{@{\extracolsep{\fill}}l*{9}{c}@{}}
\toprule
\multirow{2}{*}{\textbf{Backbone}} & \multicolumn{3}{c}{\textbf{Overall Acc.}} & \multicolumn{3}{c}{\textbf{Positive Acc.}} & \multicolumn{3}{c}{\textbf{Negative Acc.}} \\
\cmidrule(lr){2-4}\cmidrule(lr){5-7}\cmidrule(lr){8-10}
& Base & SFT & TS-Dis. & Base & SFT & TS-Dis. & Base & SFT & TS-Dis. \\
\midrule
Qwen2.5-VL-3B & 57.30 & 66.45 & 70.43 & 96.25 & 75.20 & 72.59 & 6.40 & 55.01 & 67.59 \\
Qwen2.5-VL-7B & 50.83 & 78.47 & 76.52 & 15.66 & 78.14 & 69.66 & 96.80 & 78.89 & 85.50 \\
Qwen3-VL-4B & 64.33 & 79.39 & 77.54 & 40.95 & 84.18 & 69.98 & 94.88 & 73.13 & 87.42 \\
Qwen3-VL-8B & 73.11 & 80.87 & 78.37 & 67.05 & 79.12 & 69.17 & 81.02 & 83.16 & 90.41 \\
\bottomrule
\end{tabular*}
\caption{Effects of supervised fine-tuning and teacher--student distillation on the same 1,082-transition AndroidWorld panel (\%). Inputs comprise the task, action, recent action history, and paired BEFORE/AFTER screenshots. Training is cumulative within each backbone: Base $\rightarrow$ SFT $\rightarrow$ TS-Dis.}
\label{tab:app-backbone}
\end{table}

\subsection{Refinement-order ablation}
\label{app:ablations}

We select GRPO followed by DPO using validation performance and report both
orders on the same 1,082-transition AndroidWorld test set. Both
configurations start from the same Qwen3-VL-8B
checkpoint after supervised fine-tuning and teacher--student distillation;
only the order of GRPO and DPO is changed.

\begin{table}[htbp]
\centering
\small
\begin{tabular}{@{}lccc@{}}
\toprule
\textbf{Configuration} & \textbf{Overall} & \textbf{Pos.} & \textbf{Neg.} \\
\midrule
DPO $\rightarrow$ GRPO & 80.41 & 75.04 & 87.42 \\
GRPO $\rightarrow$ DPO & 82.16 & 86.79 & 76.12 \\
\bottomrule
\end{tabular}
\caption{Overall, positive-class, and negative-class accuracy (\%) for the
refinement-order ablation on 1,082 AndroidWorld transitions.}
\label{tab:app-controlled}
\label{tab:ablations}
\end{table}

Applying GRPO before DPO yields higher overall accuracy than the reverse
order (82.16\% vs.\ 80.41\%) and improves positive-class accuracy by
11.75 percentage points (Table~\ref{tab:app-controlled}). Reversing the order
increases negative-class accuracy from 76.12\% to 87.42\%, but rejects more
valid \mbox{transitions}.

\subsection{Recovery from an ineffective scroll}
\label{app:brightness-case}

\begin{figure}[p]
\centering
\includegraphics[width=\textwidth]{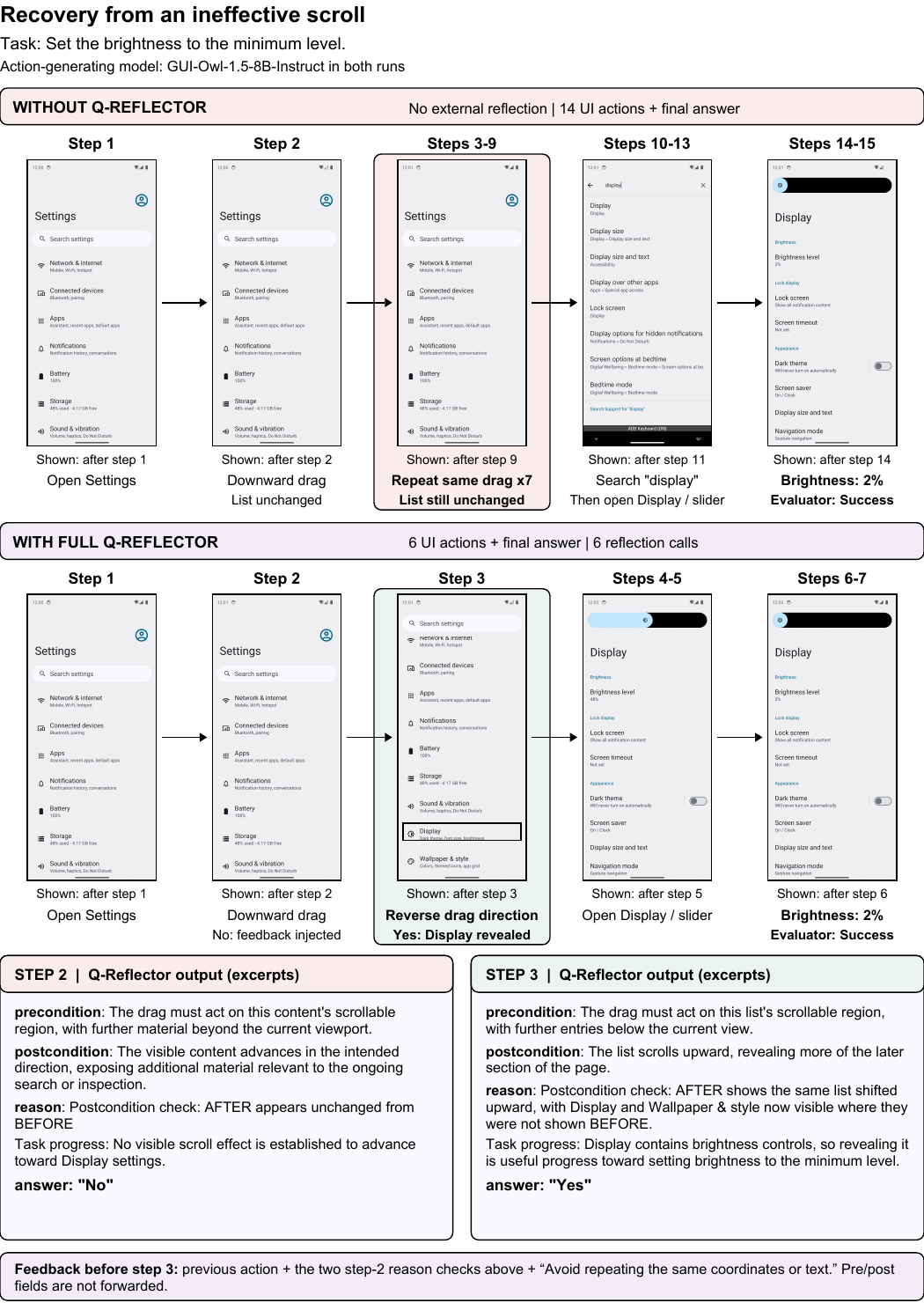}
\caption{Recorded MobileWorld trajectories without external reflection
(top) and with full \tool (bottom). Both rows use original screenshots
and step numbers; ranges compress intermediate actions. Highlighted
nodes show repetition versus recovery. Reflection excerpts are verbatim.
Both runs succeed.}
\label{fig:brightness-case}
\end{figure}

Figure~\ref{fig:brightness-case} compares two recorded runs with
GUI-Owl-1.5-8B-Instruct on the task ``Set the brightness to the minimum
level.'' Both runs open Settings at step~1 and execute the same downward
finger drag at step~2, leaving the visible list unchanged.

\paragraph{Repeated actions without reflection.}
In the upper row, the model describes its intended action as ``Scroll down
to find the Display settings'' but repeats the same ineffective gesture at
steps~3--9. The screenshots at steps~2 and~9 show that the list remains unchanged
after seven further attempts. At step~10, the model switches to search,
eventually opens Display and the brightness slider, adjusts the brightness,
and returns its final answer at step~15.

\paragraph{Detecting the failed scroll.}
In the lower row, \tool evaluates the first ineffective drag at step~2.
Its generated precondition requires a scrollable region with further content
beyond the viewport. The generated postcondition specifies the expected
local effect: ``The visible content advances in the intended direction,
exposing additional material relevant to the ongoing search or inspection.''
Comparing this expectation with the screenshots, the \emph{Postcondition
check} reports that ``AFTER appears unchanged from BEFORE''. The
\emph{Task progress} check concludes: ``No visible scroll effect is established
to advance toward Display settings.'' \tool therefore returns \texttt{No}.

\paragraph{Recovery and valid intermediate progress.}\mbox{}\\
Before step~3, the adapter forwards the previous action and these two checks,
together with ``Avoid repeating the same coordinates or text.'' The generated
condition fields remain internal to \tool. The model then drags its finger
upward, shifting the list to reveal Display, as shown in the step-3
screenshot in the lower row. \tool confirms the visible change and returns \texttt{Yes},
explaining: ``Display contains brightness controls, so revealing it is useful
progress toward setting brightness to the minimum level.'' This judgment
accepts a useful navigation step even though the brightness is still unchanged.
The model subsequently opens Display and the slider at steps~4--5, adjusts
brightness at step~6, and returns its final answer at step~7.

Both runs are marked successful by the MobileWorld evaluator. The case
illustrates how \tool identifies an ineffective action and subsequently
accepts a valid recovery step, providing a qualitative comparison of the
observed behavior.

\end{document}